\documentclass[letterpaper]{article} 

\usepackage[draft]{aaai2027}

\usepackage[hyphens]{url}  
\usepackage{graphicx} 
\usepackage{natbib}  
\usepackage{amsmath}
\usepackage{amssymb}
\usepackage{caption} 
\usepackage{algorithm}
\usepackage{algorithmic}

\usepackage{newfloat}
\usepackage{listings}
\DeclareCaptionStyle{ruled}{labelfont=normalfont,labelsep=colon,strut=off} 
\floatstyle{ruled}
\newfloat{listing}{tb}{lst}{}
\floatname{listing}{Listing}

\usepackage{booktabs}

\title{MoE Proxy Models for Low-Cost Failure  Reproduction \\ and Diagnosis in LLM RL Post-Training}
\author{
{
Yikai Wang$^{1}$\equalcontrib,
Chuansai Zhou$^{2}$\equalcontrib,
Yuhang Zhou$^{1}$,
Weiqiang Wu$^{2}$,
Cong Wu$^{2}$,
}\\
{
Yue Deng$^{1}$,
Ben Feng$^{2}$,
Mingming Zhu$^{2}$,
Beirong Zhou$^{2}$,
Zhibin Wang$^{1}$,
}\\
{
Sheng Zhong$^{1}$,
Chen Tian$^{1}$,
Wangze Zhang$^{2}$
}
}
\affiliations{
    \textsuperscript{\rm 1}State Key Laboratory for Novel Software Technology, Nanjing University, China\\
    \textsuperscript{\rm 2}Huawei, China\\
}

\begin{document}

\maketitle

\begin{abstract}
Reinforcement learning (RL) post-training of large language models (LLMs) is
computationally intensive and involves complex system pipelines with
substantial debugging overhead. In practice, factors such as framework
adaptation, numerical precision, and operator implementation can cause failures, including gradient overflow and loss divergence. Reproducing such failures directly on large models requires considerable time and computational resources. This paper systematically analyzes failures encountered during large-scale RL training on the Huawei Ascend platform, summarizes representative failure types, and identifies three model-side factors relevant to fault reproduction. Based on these factors, we propose a proxy-model construction method for low-cost fault investigation and auxiliary diagnosis. It employs structure-preserving, clustering-based expert pruning to select representative experts while retaining the model's backbone architecture, routing mechanism, and basic task capabilities. Our
experimental results show that the proxy models reduce accelerator requirements by \(50\%\)--\(87.5\%\) and achieve up to a \(33.3\times\) reduction in per-step NPU-hour cost, while preserving major training dynamics and reproducing fault responses consistent with the original models. Overall, the proxy models can serve as low-cost surrogates for fault reproduction, targeted validation, and auxiliary diagnosis in RL post-training.

\end{abstract}

\section{Introduction}
\label{sec:introduction}

Reinforcement learning (RL) post-training has become an important stage for improving the instruction-following, complex reasoning, and human-preference alignment capabilities of large language models~\cite{christiano2017deep,ziegler2019fine,ouyang2022training,guo2025deepseek}. However, compared with pre-training and supervised fine-tuning, RL post-training relies on online trajectory generation, reward feedback, policy updates, and multi-model coordination, resulting in greater stochasticity and system complexity~\cite{ouyang2022training,sheng2024hybridflow,hu2024openrlhf}. Subtle inconsistencies in framework adaptation, weight synchronization, numerical precision, operator implementation, and optimization configurations may lead to reward stagnation, abnormal KL growth, or training divergence~\cite{henderson2018deep,andrychowicz2021matters,huang2024n+,
zhong2026diagnosing,qi2025defeating}. Such faults may emerge only after prolonged training, making reproduction and validation on the original large model computationally expensive~\cite{wortsman2023small}. This challenge is particularly pronounced for Mixture-of-Experts (MoE) models, as dynamic expert routing and sparse computation further complicate the reproduction of training behavior~\cite{fedus2022switch,dai2024deepseekmoe,lo2025closer}. 
Therefore, constructing a lightweight proxy model that preserves major
training dynamics and fault-response characteristics of the original model
represents a promising approach to reducing the cost of fault analysis and diagnosis.

Existing proxy-model approaches mainly focus on efficiency optimization, hyperparameter exploration, or stability analysis during pre-training~\cite{yang2021tuning,wortsman2023small}, while model compression techniques such as pruning, quantization, and knowledge distillation primarily optimize inference efficiency, compression ratio, or task performance~\cite{frantar2023sparsegpt,frantar2022gptq,hinton2015distilling}. These methods generally do not consider whether a compressed model preserves the model-side characteristics through which post-training faults are manifested~\cite{lee2025stun,lu2024not,liu2024efficient,xie2024moe}. For MoE models, directly removing, averaging, or merging experts may alter static routing preferences, expert-utilization patterns, and hidden-state representations~\cite{lu2024not,lo2025closer,li2026sub}. Consequently, a compressed model may retain certain inference capabilities while failing to reproduce the fault-induced anomalies and training trends of the original model.

To address this limitation, we propose a multi-view, frequency-aware expert
pruning method for constructing low-cost MoE proxy models. Our method first summarizes representative issues in large-scale RL post-training~\cite{huang2024n+,zhong2026diagnosing,qi2025defeating} and identifies three fault-relevant model characteristics: routing decisions, expert-utilization patterns, and hidden-state representations. To preserve these characteristics, we model expert similarity from three perspectives: Router
parameters, expert co-activation behavior, and routed-context representations. The resulting distances are fused and used for K-Medoids clustering, through which representative experts are selected to construct the proxy model. Without parameter averaging or additional fine-tuning, the proposed method preserves the backbone architecture, Top-\(k\) routing rule, and standard MoE execution structure.

We evaluate the proposed method on the Qwen and DeepSeek families of MoE
models~\cite{yang2025qwen3,liu2025deepseek,xu2026deepseek}. The resulting proxy models reduce accelerator requirements by
\(50\%\)--\(87.5\%\) and achieve up to a \(33.3\times\) reduction in per-step NPU-hour cost. Under fault-free RL post-training, the proxy models preserve the major evolution trends of reward and actor KL loss, indicating that they maintain consistent training dynamics with the original models. Under faults related to numerical consistency and training stability, the proxy models further reproduce consistent anomaly directions and temporal trends. These results demonstrate that the proposed proxy models provide computationally efficient surrogates for low-cost fault reproduction, targeted validation, and auxiliary diagnosis.

Our main contributions are as follows:
\begin{itemize}
    \item We summarize representative fault categories, anomalous manifestations, and primary causes in large-scale RL post-training, and define a proxy-model fault-reproduction criterion based on anomaly directions and training trends.
    
    \item We propose a multi-view, frequency-aware MoE expert pruning method that constructs lightweight proxy models while preserving the backbone architecture, Top-\(k\) routing mechanism, and standard MoE execution structure.
    
    \item We evaluate the proposed proxy models across different MoE
    architectures and expert budgets, demonstrating substantial cost reductions while preserving task capability, fault-response consistency, and RL training trends for low-cost fault reproduction and auxiliary diagnosis.
\end{itemize}

The remainder of this paper is organized as follows.
Section 2 discusses RL post-training instability, proxy-model motivation, and
the limitations of existing approaches.
Section 3 presents the fault-aware proxy construction framework and the
proposed multi-view, frequency-aware expert pruning method.
Section 4 evaluates its computational efficiency, training-dynamics fidelity,
fault reproduction capability, and expert-selection effectiveness.
Finally, Section 5 concludes the paper.
\section{Background}
\label{sec:background}

\subsection{Instability of RL Post-Training} 
Reinforcement learning (RL) post-training has become a critical stage for improving the instruction-following, reasoning, and alignment capabilities of large language models~\cite{christiano2017deep,ziegler2019fine,ouyang2022training,guo2025deepseek}. However, RL is inherently characterized by high variance, non-stationarity, and sensitivity to hyperparameters~\cite{schulman2015high,henderson2018deep,andrychowicz2021matters,spangher2025rlhf}. When applied to large language models, interactions among online sampling, reward modeling, KL regularization, and complex system pipelines further amplify training instability~\cite{sheng2024hybridflow,hu2024openrlhf}. As a result, subtle inconsistencies in weight synchronization, numerical precision, operator implementation, or optimization configurations can lead to reward stagnation, abnormal KL growth, gradient overflow, or even training divergence~\cite{huang2024n+,zhong2026diagnosing,qi2025defeating}. These anomalies often emerge only after prolonged training, by which time substantial computational resources have already been consumed, while their root causes remain difficult to identify. Therefore, reproducing training anomalies and identifying their primary causes at low cost has become an urgent problem for improving the stability of RL post-training.

\subsection{Motivation for Proxy-Model Validation}
RL post-training involves multiple models (policy, reference, reward, and optional value models) and continuously performs autoregressive generation and online sampling~\cite{ouyang2022training}. Consequently, reproducing and diagnosing faults directly on the target large model is prohibitively expensive. A natural strategy to reduce this cost is to employ a lightweight proxy model—a compressed variant of the original model that can be executed on fewer accelerators while preserving sufficient structural and behavioral fidelity to serve as a surrogate for fault analysis.
This idea has shown initial promise in the pre-training stage, where small-scale models have been used for low-cost hyperparameter search and stability analysis. Representative examples include $\mu$Transfer, which enables hyperparameters tuned on small-scale models to be transferred to substantially larger models~\cite{yang2021tuning}, and small-scale Transformer proxies that reproduce large-model training instabilities and evaluate their mitigation strategies at reduced cost~\cite{wortsman2023small}. These prior findings suggest that proxy-based validation is a viable concept, motivating us to explore its systematic construction specifically for RL post-training.

\subsection{Limitations of Existing Proxy} 
Existing proxy-model approaches and model compression techniques, including pruning, knowledge distillation, and quantization, provide general-purpose foundations for constructing lightweight surrogate models~\cite{frantar2023sparsegpt,hinton2015distilling,frantar2022gptq}. However, these approaches are designed primarily to reduce model size, improve inference efficiency, or preserve task performance, and generally do not consider whether the resulting models retain the architectural structures, execution paths, and training dynamics associated with faults~\cite{lee2025stun,lu2024not,liu2024efficient,xie2024moe,
guo2025cluster,zhang2025diversifying,chen2022task}. When applied to RL post-training, such proxy models suffer from three fundamental limitations that prevent their direct use for fault diagnosis.

First, compression methods such as distillation primarily preserve final output distributions, whereas RL post-training faults often originate from intermediate execution stages, rollout generation, gradient propagation, or weight synchronization. These process-level anomalies may be smoothed over or masked by output-oriented compression objectives, making the compressed model insensitive to the very faults we aim to diagnose.

Second, RL feedback signals are sparse and delayed; a compressed model that retains only static inference capability may fail to produce valid exploration trajectories or discriminative reward distributions~\cite{roy2025you}, causing its training dynamics to deviate substantially from those of the original model. Moreover, simply scaling down model components proportionally does not preserve the conditions under which reward hacking, over-optimization, or training instability emerge~\cite{amodei2016concrete,casper2023open,
gao2023scaling,rafailov2024scaling}, as these phenomena are sensitive to the relative scale of capacity and regularization.

Third, for Mixture-of-Experts (MoE) architectures~\cite{fedus2022switch,jiang2024mixtral,dai2024deepseekmoe,
liu2024deepseek}, common practices such as expert merging or parameter averaging may alter router decision regions, Top-\(k\) expert combinations, and expert co-activation patterns~\cite{li2026sub}, all of which influence the manifestation of routing-related faults. Altering these mechanisms changes the dynamic expert-selection behavior and hidden-state propagation paths, making it unlikely that the compressed model will reproduce routing-related anomalies observed in the original model~\cite{lo2025closer}.

These limitations reveal a fundamental gap between existing proxy models and the requirements of fault diagnosis in RL post-training. Therefore, constructing a proxy model for this setting demands a specialized approach that explicitly preserves fault-relevant structures, routing dynamics, and hidden-state propagation characteristics, rather than simply optimizing for task accuracy or speed.
\section{System Design}
\label{sec:system-design}




This section presents the construction of a lightweight proxy model for low-cost fault validation and auxiliary diagnosis in RL post-training. We first discuss the key objectives and challenges of proxy-model construction and then present the proposed method.

\subsection{Design Objectives and Challenges}
\label{sec:design-goals}
The central question in proxy-model construction is: how can we minimize model size while preserving the ability to reproduce fault-related behaviors of the original model? This involves two interconnected challenges.

\textit{First, we must determine which architectural and behavioral characteristics are essential for fault reproduction.} RL post-training faults are diverse and highly coupled—similar symptoms may arise from different causes, while the same cause may manifest differently. Without a systematic identification of fault-relevant factors, pruning risks removing structures critical to fault sensitivity.

\textit{Second, we must achieve compression without destroying these characteristics.} Conventional compression methods prioritize model size, inference speed, or task performance~\cite{frantar2023sparsegpt,frantar2022gptq,hinton2015distilling}, with limited consideration of whether the compressed model retains fault-related execution paths, routing behaviors, or training dynamics.

These challenges indicate that proxy-model construction must jointly optimize for both compression efficiency and fault-response preservation, rather than minimizing parameter count alone.

\subsection{Proposed Method}
\label{sec:proposed-method}

To address the above challenges, we first characterize representative RL post-training faults and identify the model characteristics that should be preserved during proxy construction. These characteristics guide the multi-view expert-similarity design introduced in the following subsections.

\subsubsection{Problem Characterization}
\label{sec:problem-characterization}

We characterize practical RL post-training faults, identify the fault-relevant characteristics that guide expert pruning, and define the criterion for evaluating the proxy model.

Based on our experience with large-scale RL post-training on the Huawei Ascend platform, we summarize representative faults and analyze their typical manifestations and dominant causes, as shown in Table~\ref{tab:issue_taxonomy}~\cite{huang2024n+,zhong2026diagnosing,qi2025defeating}.

Although the fault categories may overlap, their dominant causes highlight three model characteristics that are particularly relevant to MoE proxy construction: routing decisions, expert-utilization patterns, and hidden-state representations. Routing decisions determine how inputs are assigned to different experts and directly affect the execution paths of MoE models~\cite{fedus2022switch,dai2024deepseekmoe,lo2025closer}. Expert-utilization patterns reflect dynamic routing relationships and expert co-activation behavior under different inputs~\cite{lo2025closer,lee2025stun}. Hidden-state representations characterize the input representations received and processed by individual experts. Accordingly, these characteristics are modeled through the Router-parameter, co-activation, and routed-context views, respectively. Preserving them during expert pruning is therefore essential for constructing a reliable proxy model.

We do not require the proxy and original models to produce numerically identical training metrics. Instead, we consider a fault reproduced when the two models exhibit similar anomaly types, metric-change directions, and temporal trends under the same fault condition.

\begin{table*}[t]
\centering
\small
\begin{tabular}{
p{0.23\linewidth}
p{0.32\linewidth}
p{0.39\linewidth}
}
\toprule
\textbf{Issue Category} &
\textbf{Typical Symptoms} &
\textbf{Primary Causes} \\
\midrule

Framework and model-state consistency issues &
Training--inference divergence; checkpoint or recovery failures &
Backend inconsistencies; weight synchronization or conversion errors; parallelism configuration mismatches \\

\addlinespace

Operator implementation and numerical consistency issues &
Numerical deviations; probability or gradient discrepancies &
Operator differences; precision errors; data-layout or reduction-order inconsistencies \\

\addlinespace

Expert routing consistency issues &
Routing shifts; expert imbalance; abnormal expert selection &
Router errors; Top-\(k\) mismatches; expert mapping or dispatch inconsistencies \\

\addlinespace

Optimization and training stability issues &
Reward degradation; KL divergence increase; training instability &
Improper optimization settings; reward or KL configuration issues; inadequate gradient clipping \\

\bottomrule
\end{tabular}
\caption{Taxonomy of representative issues in large-scale RL post-training.}
\label{tab:issue_taxonomy}
\end{table*}

\subsubsection{Proxy Model Construction}

\begin{figure*}[t]
\centering
\includegraphics[width=0.8\textwidth]{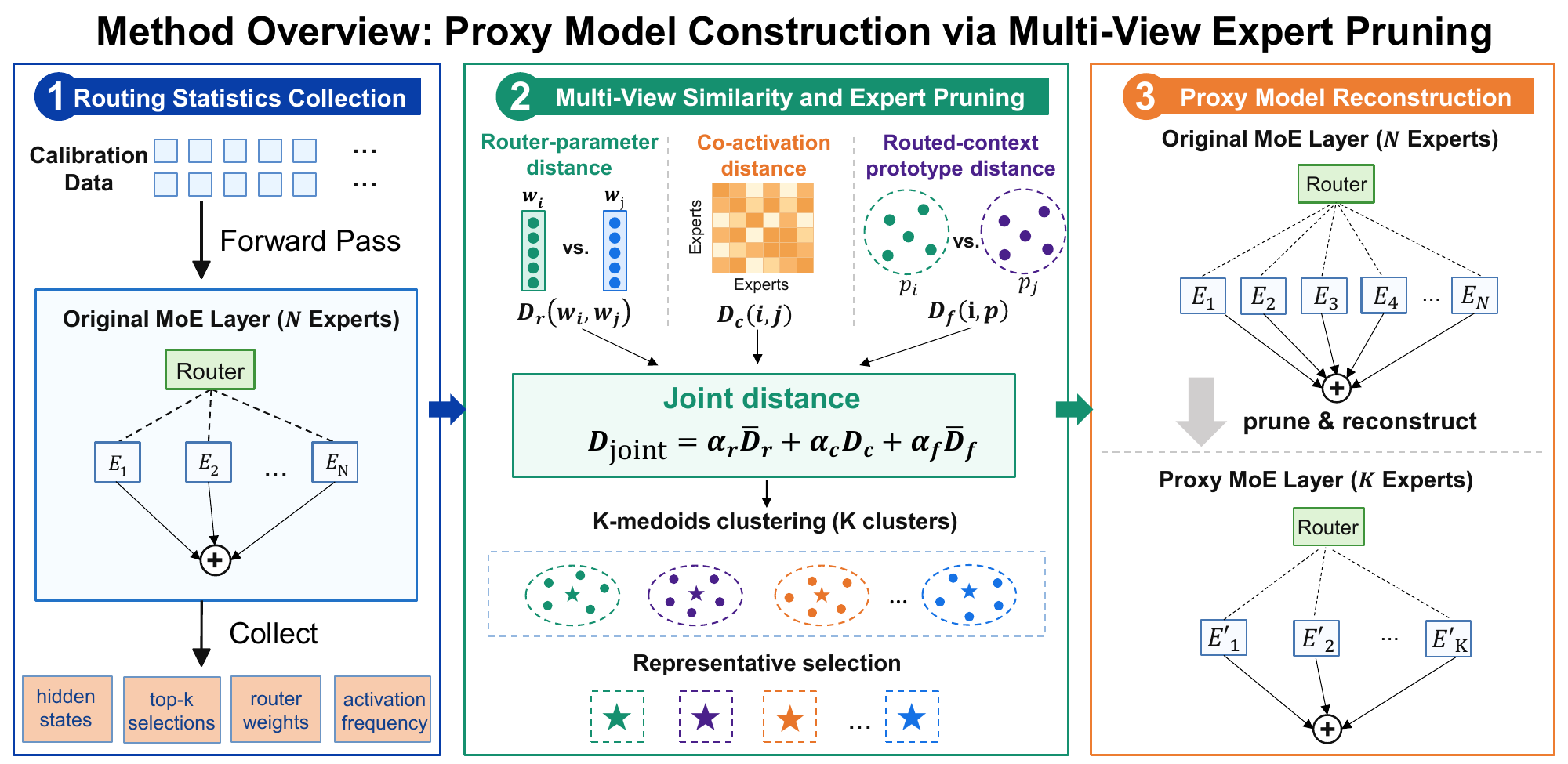} 
\caption{Overview of the proposed fault-aware proxy model construction framework via multi-view and frequency-aware expert pruning. The framework collects routing statistics from the original MoE model, measures expert similarity from router parameters, co-activation behavior, and routed-context representations, and reconstructs a lightweight proxy model by selecting representative experts while preserving the original MoE execution path.}
\label{method_workflow}
\end{figure*}

The goal is to reduce the number of MoE experts while preserving fault-relevant characteristics. We propose a multi-view, frequency-aware pruning method that selects representative experts without parameter averaging or additional fine-tuning. The overall framework is illustrated in Figure~\ref{method_workflow}, and Algorithm~\ref{alg:multi_view_expert_pruning} summarizes the procedure.

\begin{algorithm}[tb]
\caption{Multi-View and Frequency-Aware Expert Pruning}
\label{alg:multi_view_expert_pruning}
\textbf{Input}: Original MoE model $\mathcal{M}$ and calibration statistics
$\mathcal{S}_{\mathrm{cal}}$\\
\textbf{Parameter}: Target expert number $K$, fusion weights
$\alpha_{\mathrm{r}},\alpha_{\mathrm{c}},\alpha_{\mathrm{f}}$,
and routing-mass threshold $\tau_m$\\
\textbf{Output}: Proxy model $\mathcal{M}_{\mathrm{proxy}}$ and expert mappings
$\{\pi^{l}\}_{l=1}^{L}$

\begin{algorithmic}[1]

\STATE Initialize $\mathcal{M}_{\mathrm{proxy}}$ from $\mathcal{M}$.

\FOR{$l=1$ to $L$}

    \STATE Compute router distance $D_{\mathrm{r}}^{l}$.
    \STATE Compute co-activation distance $D_{\mathrm{c}}^{l}$.
    \STATE Compute routed-context distance $D_{\mathrm{f}}^{l}$.
    \STATE Normalize $D_{\mathrm{r}}^{l}$ and $D_{\mathrm{f}}^{l}$.
    \STATE Replace unreliable $\widetilde{D}_{\mathrm{f}}^{l}(i,j)$ with $\widetilde{D}_{\mathrm{r}}^{l}(i,j)$ according to $\tau_m$.
    \STATE Fuse the three distances into $D_{\mathrm{joint}}^{l}$.

    \STATE Initialize $K$ medoids using the most frequently activated experts.
    \STATE Apply K-Medoids on $D_{\mathrm{joint}}^{l}$ to obtain
    clusters $\{\mathcal{C}_{k}^{l}\}_{k=1}^{K}$.

    \FOR{$k=1$ to $K$}
        \STATE Select the most frequently activated expert
        $r_k$ from $\mathcal{C}_{k}^{l}$.
        \STATE Set $\pi^{l}(k)=r_k$.
        \STATE Copy expert $r_k$ and its router parameters into
        $\mathcal{M}_{\mathrm{proxy}}$.
    \ENDFOR

\ENDFOR

\STATE Update each MoE layer from $N$ experts to $K$ experts.
\STATE Preserve the original backbone and Top-$k$ routing mechanism.
\STATE \textbf{return}
$\mathcal{M}_{\mathrm{proxy}}$ and $\{\pi^{l}\}_{l=1}^{L}$.

\end{algorithmic}
\end{algorithm}

Consider the \(l\)-th MoE layer with \(N\) experts. Let \(\mathcal{T}\) denote the calibration set. For each token \(t\), let \(\mathbf{h}_t^l\) be the input hidden state, \(\mathcal{E}_t^l\) the Top-\(k\) selected expert set, and \(p_{t,i}^l\) the routing weight for expert \(i\). If expert \(i\) is not selected, \(p_{t,i}^l = 0\).

\noindent\textbf{Multi-View Expert Similarity.}
We measure expert differences from three complementary perspectives:

\emph{Router-Parameter Distance.} Router parameters determine the static preferences with which experts are selected for different input hidden states and are therefore a key factor affecting expert-routing consistency. We use the weight vectors \(\mathbf{w}_i^l\) to capture static routing preferences. If a bias exists, it is concatenated. The distance is:
\[
D_{r}^{l}(i,j) = \|\mathbf{w}_i^l - \mathbf{w}_j^l\|_2. \tag{1}
\]

\emph{Co-Activation Distance.} Expert co-activation relationships reflect the expert-combination patterns and utilization distributions induced by Top-\(k\) routing and are therefore important for characterizing dynamic routing consistency. This dynamic joint-selection behavior is quantified by the co-activation distance:
\[
D_{c}^{l}(i,j) = 1 - \frac{\sum_{t\in \mathcal{T}} \mathbb{I}[i\in \mathcal{E}_t^l \wedge j\in \mathcal{E}_t^l]}
{\sqrt{\sum_{t\in \mathcal{T}} \mathbb{I}[i\in \mathcal{E}_t^l] \cdot \sum_{t\in \mathcal{T}} \mathbb{I}[j\in \mathcal{E}_t^l]} + \epsilon}. \tag{2}
\]
Here, \(\mathbb{I}[\cdot]\) denotes the indicator function.

\emph{Routed-Context Prototype Distance.} The routed-context prototype distance is introduced to preserve the input representation characteristics of the hidden states routed to different experts, and is defined as:
\[
D_{f}^{l}(i,j) = \left\|
\frac{\sum_{t\in \mathcal{T}} p_{t,i}^l \mathbf{h}_t^l}{\|\sum_{t\in \mathcal{T}} p_{t,i}^l \mathbf{h}_t^l\|_2 + \epsilon}
-
\frac{\sum_{t\in \mathcal{T}} p_{t,j}^l \mathbf{h}_t^l}{\|\sum_{t\in \mathcal{T}} p_{t,j}^l \mathbf{h}_t^l\|_2 + \epsilon}
\right\|_2. \tag{3}
\]

\noindent\textbf{Distance Fusion and Proxy Reconstruction.}
To eliminate scale differences, we normalize \(D_r\) and \(D_f\) using the 95th percentile of their off-diagonal entries, clipping to \([0,1]\); \(D_c\) is already within this range. For experts with insufficient calibration support (i.e., cumulative routing mass \(m_i^l = \sum_{t} p_{t,i}^l < \tau_m\)), we replace their \(D_f\) with \(D_r\). Since Router parameters do not depend on specific activation samples, this strategy provides a more stable structural-similarity estimate for low-support experts. The joint distance is:
\[
D_{\text{joint}}^{l}(i,j) = \alpha_r \tilde{D}_r^{l}(i,j) + \alpha_c D_c^{l}(i,j) + \alpha_f \tilde{D}_f^{l}(i,j), \tag{4}
\]
where \(\alpha_r + \alpha_c + \alpha_f = 1\).

We apply K-Medoids clustering using \(D_{\text{joint}}\), initializing medoids with the most frequently activated experts. From each resulting cluster, we select the most frequently activated expert as the representative. For each MoE layer, we copy the MLP parameters and corresponding Router weights of the selected experts to reconstruct the proxy model. This preserves the original backbone and Top-\(k\) routing mechanism without introducing parameter averaging or additional fine-tuning.

\section{Experiments}
\label{sec:experiments}
This section evaluates the proposed proxy models from four perspectives:
computational efficiency, training-dynamics fidelity, fault-reproduction
capability, and expert-selection effectiveness. We first quantify the resource
savings achieved by proxy construction, then examine whether the proxy models
preserve major RL training trends and reproduce representative post-training
faults, and finally analyze the contributions of the multi-view design and
expert-selection strategy.

\subsection{Experimental Setup}
\label{sec:experimental-setup}

\noindent\textbf{Models and Hardware.}
We conduct the main experiments on Qwen3-30B-A3B~\cite{yang2025qwen3}
using a Huawei A3 server equipped with 16 Ascend 910 NPUs. The original
model contains 128 routed experts in each MoE layer with Top-\(8\) routing. For Qwen3-30B-A3B, we construct proxy models retaining 48 and 64 experts per layer. We further evaluate the proposed method on DeepSeek models~\cite{liu2025deepseek} with different model scales and MoE configurations to verify its general applicability.

\noindent\textbf{Data and Training Configuration.}
We use an independent calibration set to collect expert activation frequencies, Top-\(k\) selections, routing weights, and routed hidden states. The calibration samples do not overlap with any test samples. GSM8K~\cite{cobbe2021training} is used for task-capability evaluation and as
the task environment for fault-reproduction experiments. RL post-training is implemented using VERL~\cite{sheng2024hybridflow} and vLLM~\cite{kwon2023efficient} with GRPO~\cite{shao2024deepseekmath}. Unless otherwise specified, the original and proxy models use the same RL training configuration.

\noindent\textbf{Baselines and Evaluation Metrics.}
The capability comparisons include pairwise-view ablations, frequency-only
selection, random selection, and fixed First-\(K\) expert selection. For fault
reproduction, we compare our method with frequency-only selection and
model-centric expert grouping and merging~\cite{li2026sub}. We inject two representative faults: the Rollout LogP Precision Fault for numerical consistency and the Actor Update Omission Fault for optimization stability. The evaluation metrics include task accuracy, computational cost, rollout--training LogP discrepancy, KL divergence, and training-dynamics consistency.

\subsection{Computational Efficiency}
\label{sec:computational-efficiency}

We first evaluate the reduction in RL post-training resource costs achieved by the proxy models. As shown in Table~\ref{tab:computational-efficiency}, the proxy models substantially reduce the minimum deployment scale and per-step runtime, with the resulting NPU-hour costs analyzed below.

\begin{table}[t]
\centering
\small
\begin{tabular}{lcccc}
  \toprule
  \textbf{Model} &
  \textbf{Variant} &
  \textbf{Experts} &
  \textbf{NPUs} &
  \textbf{Time/Step} \\
  \midrule
  Qwen3-30B-A3B
  & Original
  & 128
  & 16
  & 128.93 s \\

  Qwen3-30B-A3B
  & Proxy
  & 48
  & 8
  & 114.83 s \\

  DeepSeek-V3.2
  & Original
  & 256
  & 512
  & 50 min \\

  DeepSeek-V3.2
  & Proxy
  & 16
  & 64
  & 12 min \\
  \bottomrule
\end{tabular}
\caption{Computational cost of the original and proxy models during RL post-training.}
\label{tab:computational-efficiency}
\end{table}

For Qwen3-30B-A3B, the original model requires 16 Ascend 910 NPUs to launch training, whereas the proxy model retaining 48 experts requires only 8 NPUs. Under the same training pipeline, the per-step training time decreases from 128.93 seconds to 114.83 seconds, while the corresponding per-step NPU-hours decrease from 0.573 to 0.255, representing an approximately 55.5\% reduction in computational cost.

In the larger-scale DeepSeek-V3.2 experiment, the original model requires 32 nodes with a total of 512 NPUs and takes approximately 50 minutes per training step. The proxy model requires only 4 nodes with a total of 64 NPUs, reducing the per-step training time to approximately 12 minutes. Its per-step computational cost decreases from 426.7 NPU-hours to 12.8 NPU-hours, corresponding to an approximately \(33.3\times\) cost reduction.

These results demonstrate that the constructed proxy models substantially reduce the hardware requirements and computational cost of RL post-training, providing an efficient alternative for fault reproduction, validation, and diagnostic analysis.

\subsection{Training-Dynamics Fidelity}
\label{sec:training-dynamics}

Beyond reducing computational cost, an effective proxy model should preserve
the major optimization dynamics of the original model during RL post-training. To evaluate this property, we compare the reward and actor KL loss of the original
DeepSeek-V4-Flash model~\cite{xu2026deepseek} with those of its 64-expert and
32-expert proxy models.

\begin{figure}[t]
\centering
\includegraphics[width=0.95\columnwidth]{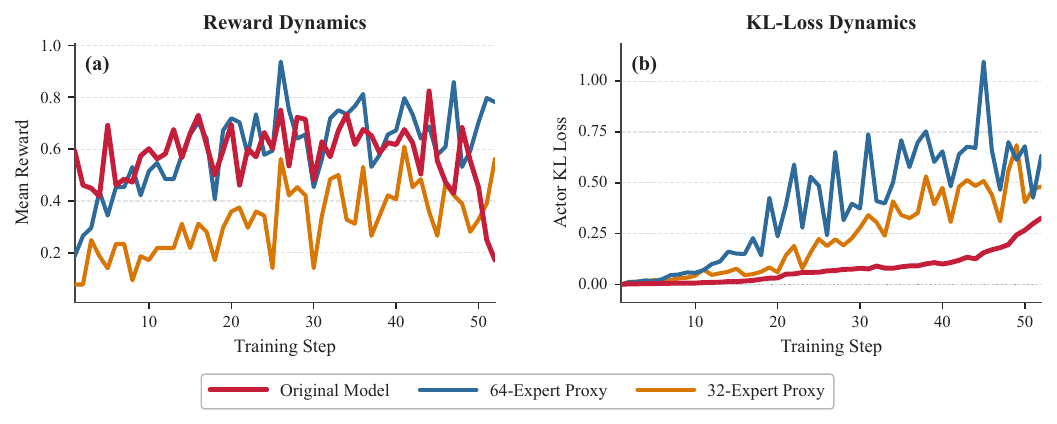}
\caption{
Reward and actor KL-loss dynamics of the original DeepSeek-V4-Flash model
and its 64-expert and 32-expert proxy models. The proxy models preserve the
major training trends of reward improvement and KL-loss growth, while retaining
more experts reduces the deviation from the original model.
}
\label{fig:training-dynamics}
\end{figure}

As shown in Figure~\ref{fig:training-dynamics}(a), all three models exhibit an
overall reward improvement trend despite short-term fluctuations. The
64-expert proxy model achieves a reward trajectory closer to the original
model, while the 32-expert proxy model preserves the same increasing trend
with a larger performance gap. These results indicate that the proxy models preserve the overall reward evolution of the original model, while retaining more experts improves training-trajectory fidelity.

Figure~\ref{fig:training-dynamics}(b) shows that the actor KL loss of all
models gradually increases during training, reflecting a consistent evolution
of policy distribution shifts. Although the proxy models differ from the
original model in absolute KL magnitude and local fluctuations, their overall
change directions remain consistent.

\subsection{Fault Reproduction Capability}
\label{sec:fault-reproduction}

After verifying that the proxy models preserve the major RL training dynamics, we further evaluate whether they can reproduce the fault-induced behaviors of the original models. A fault is considered reproduced when the proxy and original models exhibit consistent anomaly types, metric-change directions, and temporal trends under the same fault condition. Based on the issue categories summarized in Table~\ref{tab:issue_taxonomy}, we select two representative faults for evaluation: the Rollout LogP Precision Fault, which reflects numerical consistency issues, and the Actor Update Omission Fault, which reflects training stability issues during optimization.

\noindent\textbf{Rollout LogP Precision Fault.} This fault introduces a probability discrepancy between the generation and training stages by reducing the numerical precision of the scaling factor used in Rollout LogP computation. Except for whether the fault is enabled, the healthy and faulty runs use identical data and training configurations.

\begin{figure}[t]
\centering
\includegraphics[width=0.95\columnwidth]{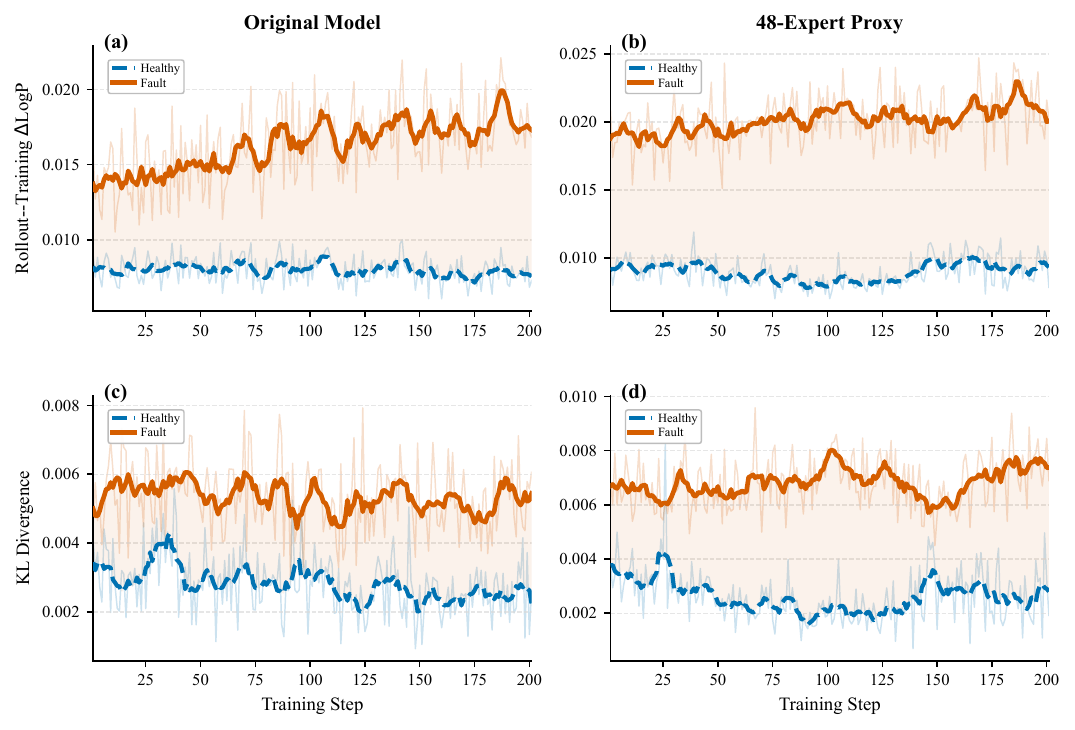}
\caption{
Comparison of Rollout--Training LogP discrepancy and KL divergence under the Rollout LogP Precision Fault. The original model and 48-expert proxy model exhibit consistent anomaly patterns after fault injection, including LogP shift and KL divergence increase.
}
\label{fig:logp-fault}
\end{figure}

Figure~\ref{fig:logp-fault} compares the Rollout--Training LogP discrepancy and KL divergence of the original model and the 48-expert proxy model under healthy and faulty conditions. After fault injection, both models exhibit consistent anomaly patterns, including a shift in the Rollout--Training LogP discrepancy and an increase in KL divergence. Although the magnitude of the changes differs between the two models, their anomaly directions and temporal trends remain consistent, indicating that the proxy model preserves sensitivity to the numerical fault.

To further analyze the impact of proxy construction strategies on fault
reproduction, we compare our method with frequency-only selection and expert
grouping and merging. Under the same 48-expert budget, both our method and
frequency-only selection reproduce the qualitative fault signatures, including the LogP-discrepancy shift and KL-divergence increase, whereas the
expert-merging baseline fails to sustain stable training. Nevertheless, the
ability to reproduce a single fault does not alone determine the quality of a proxy model. Compared with frequency-only selection, our method additionally accounts for complementary expert relationships and achieves stronger task-capability preservation while maintaining fault sensitivity. Detailed fault-reproduction comparisons are provided in appendix.

\noindent\textbf{Actor Update Omission Fault.}
To evaluate whether the proxy model preserves sensitivity to optimization failures, we inject an Actor Update Omission Fault by disabling actor parameter updates while keeping the remaining RL pipeline unchanged. Under this fault, trajectories and rewards are still collected, but the policy cannot update according to the
optimization signals, preventing further policy improvement.

\begin{figure}[t]
\centering
\includegraphics[width=0.98\columnwidth]{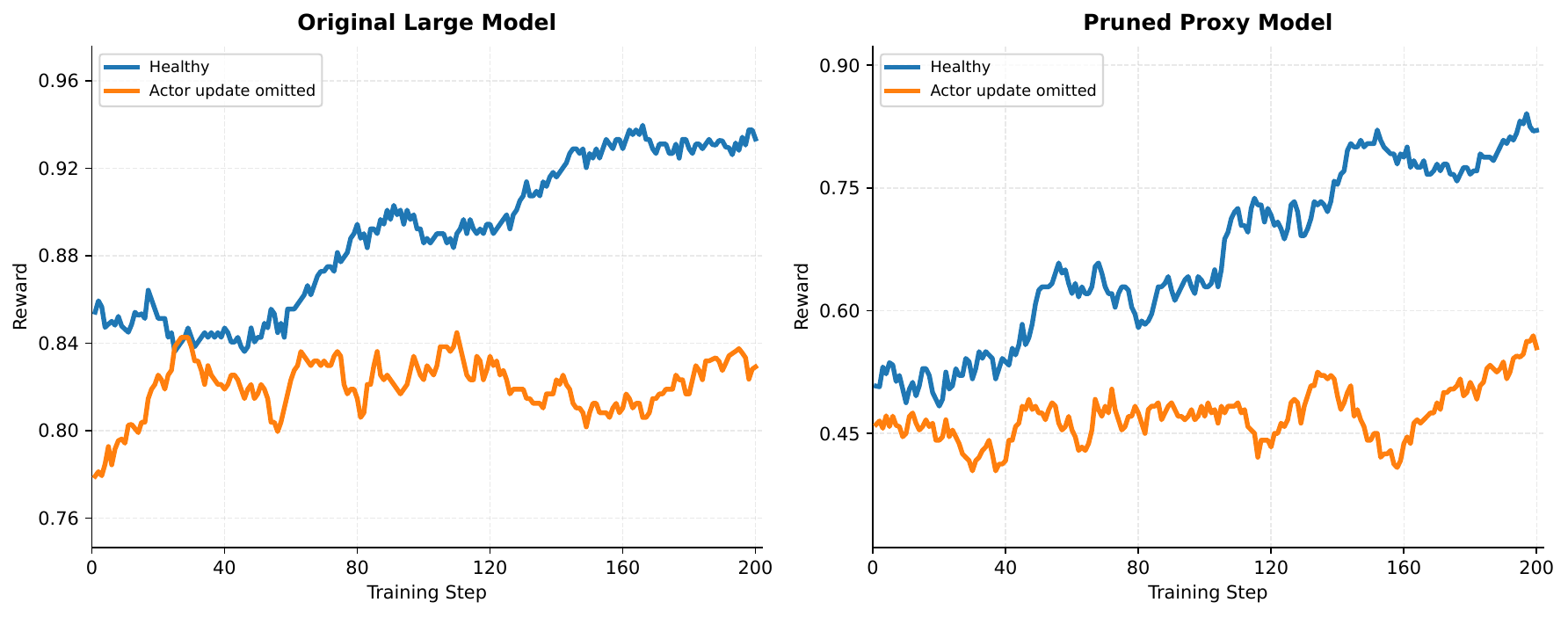}
\caption{
Reward dynamics under the Actor Update Omission Fault.
Disabling actor updates suppresses reward improvement in both the original
model and the pruned proxy model, demonstrating that the proxy model preserves
the reward-stagnation behavior caused by this optimization fault.
}
\label{fig:actor-update-omission}
\end{figure}

Figure~\ref{fig:actor-update-omission} compares the reward trajectories of the original and proxy models under healthy and faulty conditions. Since the per-step rewards are discrete and highly fluctuating, centered rolling means are used to highlight long-term trends. The runs are aligned to their common training horizon, with a wider smoothing window used for the original model. Under healthy training, both models exhibit clear reward improvement, whereas omitting actor updates substantially suppresses this trend. Although the magnitude of stagnation differs, both models exhibit the same fault-induced suppression of reward improvement, indicating that the proxy model preserves sensitivity to this optimization fault. 

\subsection{Ablation and Expert Selection Analysis}
\label{sec:capability-ablation}

We analyze the proposed expert selection method from two aspects: the contribution of different expert-similarity views and its performance relative to alternative selection strategies. We conduct the multi-view ablation on Qwen3-30B-A3B and compare expert-selection strategies on DeepSeek-V3.2.

\subsubsection{Multi-View Ablation}

Table~\ref{tab:qwen-ablation} presents the results under different expert budgets and similarity configurations. The complete three-view method achieves accuracies of 48.0\% and 86.0\% when retaining 48 and 64 experts, respectively, outperforming all pairwise-view combinations. These results indicate that the three similarity views capture complementary expert characteristics and jointly provide more effective guidance for expert selection. Increasing the number of retained experts further mitigates the capability degradation caused by pruning.

\begin{table}[t]
\centering
\small
\begin{tabular}{lcc}
  \toprule
  \textbf{Expert Similarity} &
  \textbf{48 Experts} &
  \textbf{64 Experts} \\
  \midrule
  Original Model & \multicolumn{2}{c}{93.5\%} \\
  \midrule
  Context + Co-activation & 44.0\% & 76.0\% \\
  Router + Co-activation & 46.5\% & 81.0\% \\
  Context + Router & 44.5\% & 77.5\% \\
  \textbf{Ours (Three Views)} & \textbf{48.0\%} & \textbf{86.0\%} \\
  \bottomrule
\end{tabular}
\caption{
GSM8K accuracy of Qwen3-30B-A3B proxy models under different
expert-similarity configurations.
}
\label{tab:qwen-ablation}
\end{table}

\subsubsection{Expert Selection Comparison}
\label{sec:expert-selection}

To further evaluate the effectiveness of the proposed expert selection
strategy, we compare it with frequency-only selection, random selection,
and fixed First-\(K\) expert retention on DeepSeek-V3.2.

Table~\ref{tab:deepseek-selection} compares different expert-selection
strategies when retaining 16 routed experts per layer. The proposed method
achieves an accuracy of 58.5\%, outperforming frequency-only selection by
5.5 percentage points. In contrast, random selection and fixed retention of
the first 16 experts both achieve only 3.0\%, demonstrating that pruning
experts without considering expert relationships leads to severe capability
degradation.

These results demonstrate that the proposed expert selection strategy achieves better capability preservation under aggressive expert reduction.

\begin{table}[t]
\centering
\small
\begin{tabular}{lc}
  \toprule
  \textbf{Selection Strategy} &
  \textbf{GSM8K Accuracy} \\
  \midrule
  Original Model & 95.0\% \\
  Frequency-Only & 53.0\% \\
  Random Selection & 3.0\% \\
  Fixed First-\(K\) Experts & 3.0\% \\
  \textbf{Ours} & \textbf{58.5\%} \\
  \bottomrule
\end{tabular}
\caption{
GSM8K accuracy of different expert-selection strategies on DeepSeek-V3.2.
}
\label{tab:deepseek-selection}
\end{table}

Overall, the proposed proxy models substantially reduce RL post-training cost
while retaining sufficient task capability, fault-response consistency, and major training trends. The ablation results further validate the effectiveness of the multi-view similarity design and frequency-aware expert selection.

\section{Conclusion}
\label{sec:conclusion}

This paper presents a multi-view, frequency-aware proxy-model construction method for supporting low-cost issue localization and auxiliary diagnosis in the RL post-training of Mixture-of-Experts large language models. By jointly modeling Router parameters, expert co-activation behavior, and routed-context representations, the proposed method selects representative experts while preserving the original backbone, Top-\(k\) routing mechanism, and standard MoE execution path. Experiments on Qwen and DeepSeek models demonstrate that the resulting proxy models substantially reduce hardware requirements and training costs while retaining sufficient task capability for valid RL post-training. They also exhibit consistent fault responses and major training dynamics with the original models. These results demonstrate that the proposed proxy models can serve as efficient surrogates for low-cost fault reproduction, validation, and auxiliary diagnosis before deploying expensive verification on the original models.


\bibliography{aaai2027}



\end{document}